\documentclass{article}
\usepackage[preprint]{spconf}
\usepackage{spconf,amsmath,graphicx}
\usepackage{url}
\usepackage{amsmath}
\usepackage{amssymb}
\usepackage{multirow, multicol}
\usepackage{hyperref}
\usepackage[table,xcdraw,dvipsnames]{xcolor}
\usepackage{graphicx}

\title{LS-HDIB: A Large Scale Handwritten Document Image Binarization Dataset}

\name{Kaustubh Sadekar \qquad Ashish Tiwari \qquad Prajwal Singh \qquad Shanmuganathan Raman}
  
\address{CVIG Lab, Indian Institute of Technology Gandhinagar \\ \{ sadelkar.k, ashish.tiwari, singh\_prajwal, shanmuga\}@iitgn.ac.in}
\copyrightnotice{\copyright\ IEEE 2022}

\begin{document}
\ninept
\maketitle
\begin{abstract}
Handwritten document image binarization is challenging due to high variability in the written content and complex background attributes such as page style, paper quality, stains, shadow gradients, and non-uniform illumination. While the traditional thresholding methods do not effectively generalize on such challenging real-world scenarios, deep learning-based methods have performed relatively well when provided with sufficient training data. However, the existing datasets are limited in size and diversity. This work proposes LS-HDIB - a large-scale handwritten document image binarization dataset containing over a million document images that span numerous real-world scenarios. Additionally, we introduce a novel technique that uses a combination of adaptive thresholding and seamless cloning methods to create the dataset with accurate ground truths. Through an extensive quantitative and qualitative evaluation over eight different deep learning based models, we demonstrate the enhancement in the performance of these models when trained on the LS-HDIB dataset and tested on unseen images. 

\end{abstract}

\section{Introduction}
\label{sec:intro}
\footnotetext[1]{This work is supported by SERB IMPRINT-2 grant.}
Handwritten document image binarization is generally modeled as a classification problem in which intra-image pixels are assigned to either of the two classes: handwritten content (the foreground) or the background. Document image binarization has been an active research area for decades owing to its importance as an essential pre-processing step in facilitating several document image processing tasks such as optical character recognition, handwriting matching, document translation, document summarization, and changing the background.
\begin{figure}[h]
    \centering
    \includegraphics[width=\linewidth]{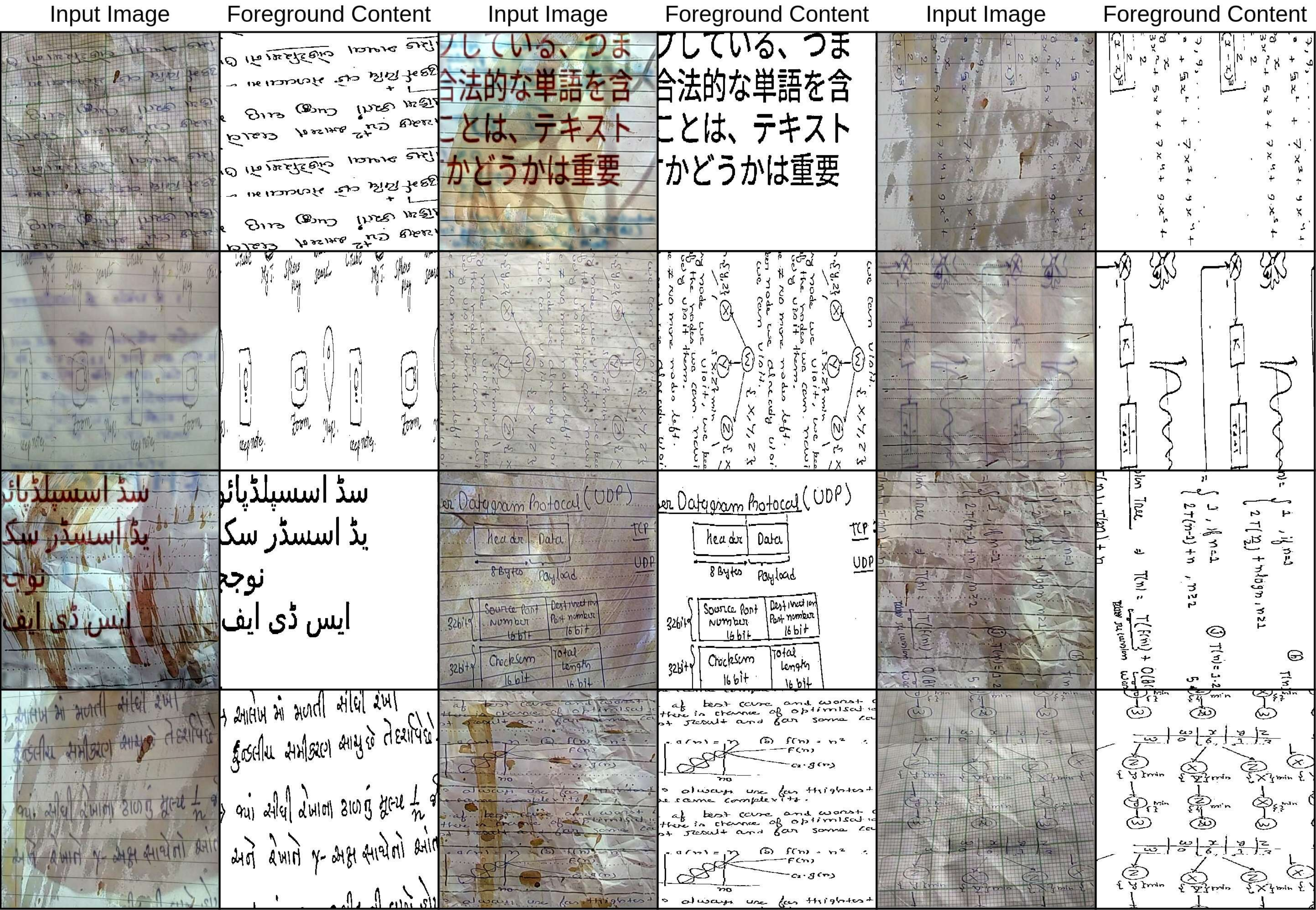}
    \caption{A few samples of handwritten document images obtained from the proposed LS-HDIB dataset.}
    \label{fig:teaser}
    \vspace{-4mm}
\end{figure}
Handwritten documents range from ancient documents, old legal records, and ledgers to music scores and handwritten bills. These documents often degrade over time and become difficult to comprehend. Common degradation scenarios include crumpled pages, poor foreground-background contrast, stains, paper aging, fainted characters, uneven illumination, and bleed/show-through. Furthermore, handwritten documents posses various page styles, i.e., grids, lines, staff annotation styles, and partially blank pages that increase the difficulty of segmenting the foreground content. The variability in the type and the thickness of strokes also increases the complexity. These challenges make handwritten document image binarization extremely difficult. Traditional methods like Otsu image segmentation \cite{otsu} and adaptive thresholding \cite{adaptivethresh} fail to address the aforementioned challenges completely. Since these algorithms utilize only the low-level features, they fail to capture the wide range of variability inherent in the handwritten documents, limiting their ability to distinguish the background from the foreground content. The high-level features can differentiate text pixels from background noises handling the degradations better. However, using them solely can cause loss of low-level information like character edges and contours, making it insufficient to address the binarization problem. Thus far, deep learning methods have shown promising results in segmenting foreground and background content \cite{Tensmeyer2017DocumentIB,deepsn} by incorporating both low-level and high-level image features. Deep learning models rely heavily on a large amount of data for better generalization \cite{robograsp}. However, document image binarization lacks such a large and diverse dataset that covers numerous real-world scenarios. While the focus has been on designing new robust networks, little attention has been given to scale up the existing datasets. 

To address the aforementioned challenges and to generate a large scalable dataset, we propose a Large Scale Handwritten Document Image Binarization dataset (LS-HDIB) that contains over a million handwritten document images. We propose a simple and effective method for generating the LS-HDIB dataset with accurate ground truths containing segmented handwritten content (see Fig. \ref{fig:teaser}). Interestingly, the proposed method requires no manual intervention for generating these segmented ground truths. The primary \textbf{contributions} of this work are: (i) A large scale dataset (LS-HDIB) containing over a million images with accurate ground truths for handwritten document image binarization. (ii)  A scalable and efficient method to generate and extend the proposed dataset.

\section{Related Work} \label{sec:related_work}

The standard approaches for document image binarization are classified into (i) \textit{global methods}\cite{otsu, tsai1985moment} which use a single threshold value and (ii) \textit{local methods} \cite{gatos2006adaptive} which use adaptive threshold values for separating the foreground and the background content. While the global methods fail to handle complex degradations, the local methods are computationally expensive and driven mainly by manual parameter tuning. Several deep neural network architectures have been designed for document image binarization \cite{pastor2015insights, calvo2017pixel, afzal2015document}. Researchers have used fully convolutional neural networks \cite{tensmeyer2019generating, peng2017using}, recurrent neural networks \cite{westphal2018document}, encoder-decoder frameworks \cite{calvo2017pixel,peng2017using}, and generative adversarial networks \cite{tensmeyer2019generating, improvegan} to address document binarization. While the performance of these learning-based frameworks depends on the robustness of the network design, another aspect critical to their performance is the size and the versatility of the training data. The existing datasets \cite{dibco9,hdibco10,dibco11,hdibco12,dibco13,hdibco14,hdibco16,dibco17,dibco18,phdib} do not completely span complex degradations, page styles, and illumination variations. Owing to a very large space spanned by the possible handwritten content and background variations, there is a need to develop a method to create a large and diverse dataset for handwritten document image binarization. In this work, we propose a simple yet effective method to create a large-sized dataset that can potentially circumvent the aforementioned limitations.

\section{Method}
\subsection{Dataset Generation} 
\begin{figure}[h]
    \centering
    \includegraphics[width=\linewidth]{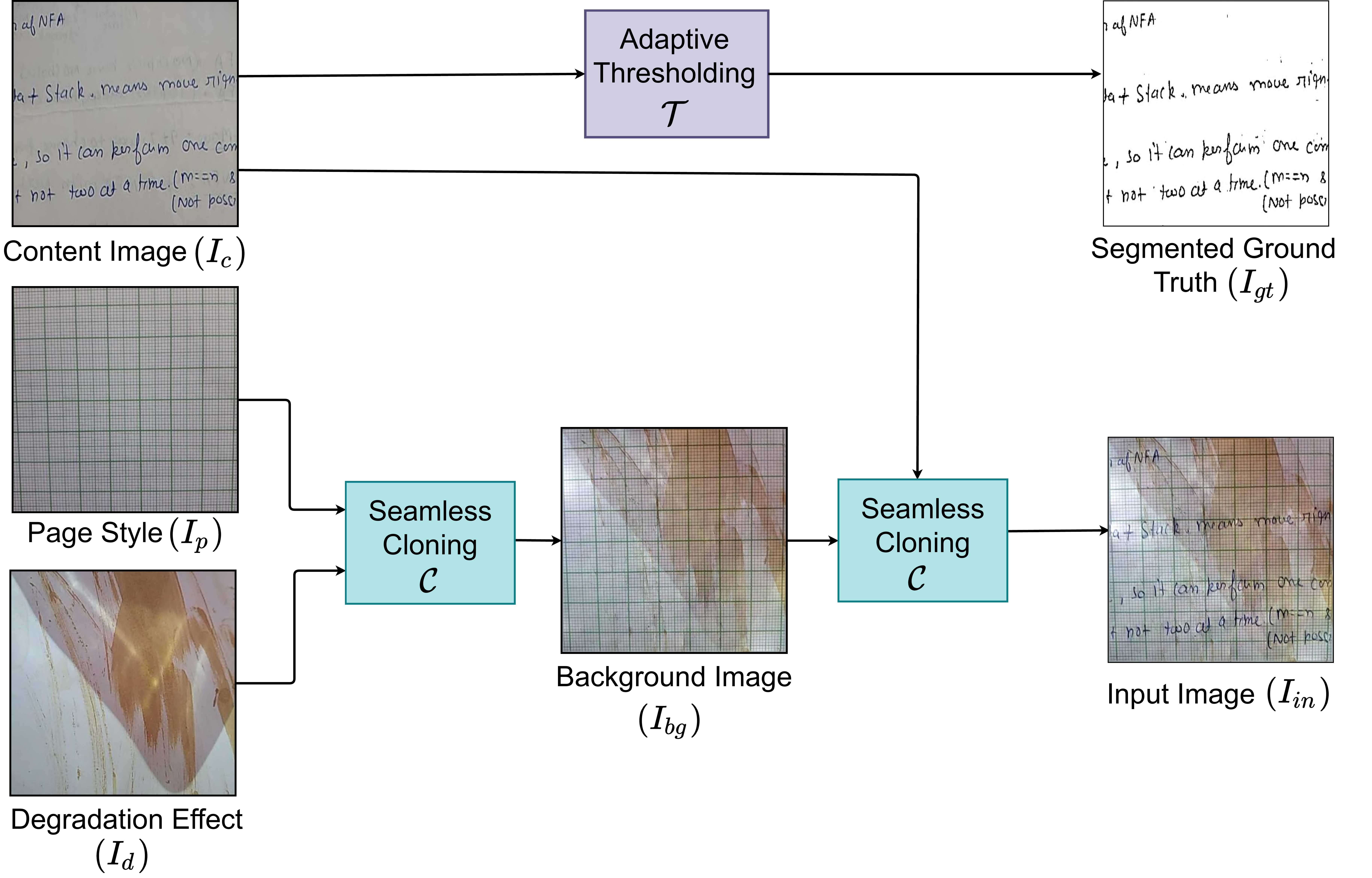}
    \caption{Block schematic of the proposed method for generating LS-HDIB dataset.}
    \label{fig:bd}
    \vspace{-4mm}
\end{figure}

We propose a novel data generation technique that uses a combination of adaptive thresholding \cite{adaptivethresh} and mixed gradient seamless cloning \cite{blending}, as described in Fig. \ref{fig:bd}. We collect the images of a variety of handwritten content over a plain background and refer to these images as \textit{full-length document images $(I_{doc})$}. We collected over $450$ full-length document images. Around $400$ images contain handwritten content in various forms such as alpha-numeric characters, electrical circuit diagrams, control system schematics, chemical molecular structures, and flow charts that are not present in the existing datasets. Further, to diversify the types, thickness, and styles of strokes, we obtained these images from $21$ different persons. Around $50$ document images were digitally created, using Google Translate, in $13$ different languages, including English, Urdu, Mandarin, Portuguese, Russian, French, Hindi, Telugu, Malayalam, Punjabi, Gujarati, Japanese, and Korean in $21$ different font styles of various sizes and colors.
\label{sec:method}

\begin{figure*}[h]
    \centering
    \includegraphics[width=\textwidth]{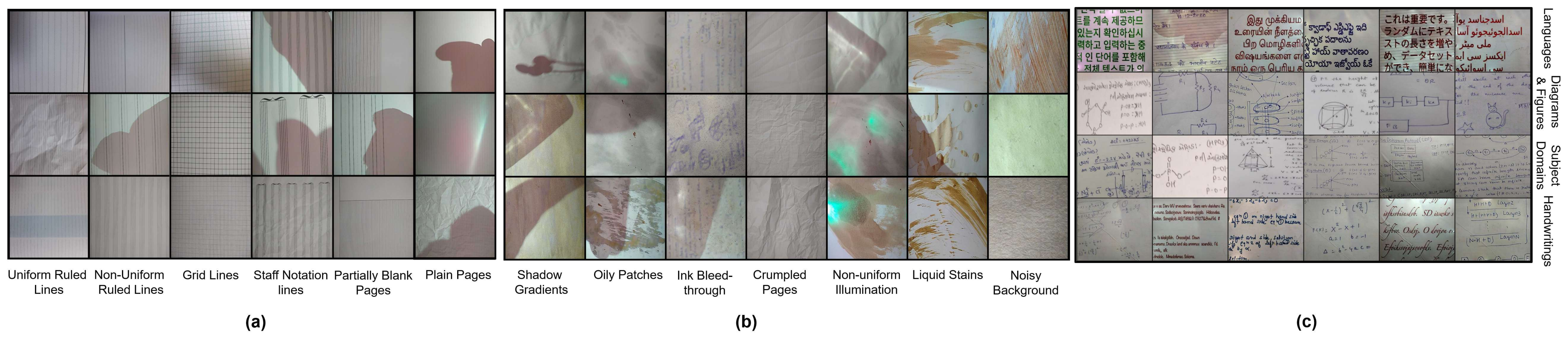}
    \caption{A few sample images depicting different page styles available in LS-HDIB dataset.}
    \label{fig:pgs}
    \vspace{-4mm}
\end{figure*}

We apply the \textit{adaptive thresholding $(\mathcal{T})$} \cite{adaptivethresh} on $(I_{doc})$  to obtain the \textit{segmented ground truth images $(I_{gt})$} such that $I_{gt} = \mathcal{T}(I_{ref})$. Next, we generate a total of $10,944$ unique \textit{content images $(I_{c})$} by cropping and augmenting multiple patches of size $480 \times 480$ from the full-length document images $(I_{doc})$. Figure \ref{fig:pgs}(a) shows a few sample images containing a variety of written content in different languages and font styles, including diagrams and texts from different subject domains. While the written content (foreground) associated with the generated ground truth remains unchanged, the background can vary depending on different page styles and degradation scenarios. We obtain multiple document images by merely changing the background with essentially the same ground truth. Moreover, this method can automatically generate ground truths saving hours of tedious manual annotation. To generate different backgrounds, we manually capture pages with a wide variety of \textit{page styles $(I_{p})$} and \textit{degradation effects $(I_{d})$} that are predominant in the real-world. We then use mixed gradient-based seamless cloning $(\mathcal{C})$ \cite{blending} to blend multiple patches of $(I_{p})$ and $(I_{d})$ to generate $20,484$ photorealistic \textit{background images $(I_{bg})$} such that $I_{bg} = \mathcal{C}(I_{p}, I_{d})$.

\begin{table*}[t]
\resizebox{\textwidth}{!}{%
\begin{tabular}{l|ccccccccccccccc}
\hline
\multicolumn{1}{c|}{\multirow{2}{*}{\textbf{Datasets}}} & \multicolumn{6}{c|}{\textbf{Page Styles}} & \multicolumn{9}{c}{\textbf{Degradation Effects}} \\ \cline{2-16}  
\multicolumn{1}{c|}{} & \begin{tabular}[c]{@{}c@{}}Uniform \\ ruled lines\end{tabular} & \begin{tabular}[c]{@{}c@{}}Non-uniform \\ ruled lines\end{tabular} & \begin{tabular}[c]{@{}c@{}}Grid \\ lines\end{tabular} & \begin{tabular}[c]{@{}c@{}}Staff notation\\ lines\end{tabular} & \begin{tabular}[c]{@{}c@{}}Partially \\ blank pages\end{tabular} & \multicolumn{1}{c|}{Plain page} & \begin{tabular}[c]{@{}c@{}}Shadow \\ gradients\end{tabular} & \begin{tabular}[c]{@{}c@{}}Oily\\ patches\end{tabular} & \begin{tabular}[c]{@{}c@{}}Ink \\ bleed-through\end{tabular} & \begin{tabular}[c]{@{}c@{}}crumpled \\ pages\end{tabular} & \begin{tabular}[c]{@{}c@{}}Non-uniform \\ illumination\end{tabular} & \begin{tabular}[c]{@{}c@{}}Noisy \\ background\end{tabular} & \multicolumn{1}{l}{\begin{tabular}[c]{@{}l@{}}Liquid \\ stains\end{tabular}} & \begin{tabular}[c]{@{}c@{}}Poor foreground-\\ background contrast\end{tabular} & \begin{tabular}[c]{@{}c@{}}Punched, stapled \\ or torn pages\end{tabular} \\ \hline
DIBCO09 & x & x & x & x & \checkmark & \multicolumn{1}{c|}{\checkmark} & x & x & \checkmark & x & x & x & \checkmark & \checkmark & x \\
HDIBCO10 & \checkmark & x & x & x & x & \multicolumn{1}{c|}{\checkmark} & x & \checkmark & \checkmark & x & x & \checkmark & \checkmark & \checkmark & x \\
DIBCO11 & x & x & x & x & x & \multicolumn{1}{c|}{\checkmark} & x & x & \checkmark & x & x & \checkmark & \checkmark & \checkmark & x \\
HDIBCO12 & \checkmark & x & x & x & x & \multicolumn{1}{c|}{\checkmark} & x & \checkmark & \checkmark & x & \checkmark & \checkmark & \checkmark & \checkmark & x \\
DIBCO13 & x & x & x & x & x & \multicolumn{1}{c|}{\checkmark} & x & \checkmark & \checkmark & \checkmark & x & \checkmark & \checkmark & \checkmark & x \\
HDIBCO14 & x & x & x & x & x & \multicolumn{1}{c|}{\checkmark} & x & x & \checkmark & x & x & x & \checkmark & \checkmark & x \\
PHIBD12 & \checkmark & x & x & x & x & \multicolumn{1}{c|}{\checkmark} & x & \checkmark & \checkmark & x & x & \checkmark & \checkmark & \checkmark & \checkmark \\
 HDIBCO16 & x & x & x & x & x & \multicolumn{1}{c|}{\checkmark} & x & \checkmark & \checkmark & x & \checkmark & \checkmark & \checkmark & \checkmark & x \\
DIBCO17 & x & x & x & x & x & \multicolumn{1}{c|}{\checkmark} & x & \checkmark & \checkmark & x & x & \checkmark & \checkmark & x & x \\
DIBCO18 & x & x & x & x & x & \multicolumn{1}{c|}{\checkmark} & x & \checkmark & \checkmark & x & x & \checkmark & \checkmark & \checkmark & x \\
Bickley Diary & x & x & x & x & \checkmark & \multicolumn{1}{c|}{\checkmark} & x & \checkmark & x & x & x & \checkmark & \checkmark & \checkmark & x \\
Palm Leaf Manuscript & x & x & x & x & \checkmark & \multicolumn{1}{c|}{\checkmark} & \checkmark & \checkmark & x & x & x & x & \checkmark & \checkmark & \checkmark \\ \hline
\textbf{LS-HDIB (Ours)} & \checkmark & \checkmark & \checkmark & \checkmark & \checkmark & \multicolumn{1}{c|}{\checkmark} & \checkmark & \checkmark & \checkmark & \checkmark & \checkmark & \checkmark & \checkmark & \checkmark & x \\\hline
\end{tabular}}

\caption{Comparison of page styles and degradation effects available across different publicly available datasets and LS-HDIB dataset.}
\label{table:dataset_summary}
\end{table*}

For a better visual understanding, we show multiple example images of different page styles and degradation effects in Fig. \ref{fig:pgs}(a) and Fig. \ref{fig:pgs}(b), respectively. Finally, we combine the content images $(I_{c})$ and the background images $(I_{bg})$ again using seamless cloning $(\mathcal{C})$ to generate the handwritten document images $(I_{in})$ such that $I_{in} = \mathcal{C}(I_{c}, I_{bg})$. To generate the LS-HDIB dataset, we randomly sample $100$ background images $(I_{bg})$ for each of the $10,994$ handwritten content images $(I_{c})$. In this way, we obtain a total of $1.09$ million images in the LS-HDIB dataset. It is important to note that we can scale up the dataset size for up to $200$ times by considering all the background images instead of just $100$. With adaptive thresholding and mixed gradient-based seamless cloning, we have been able to generate accurate ground truths without the requirement of any manual annotation and generate a wide variety of degraded handwritten document images. In addition to the inherent scalability and ease of ground truth generation, the proposed dataset generation method is computationally less expensive compared to the other deep-learning-based generative methods such as the one proposed in \cite{improvegan}. As evident from Table \ref{table:dataset_summary}, most of the publicly available datasets contain the written content only over plain pages. In contrast, the LS-HDIB dataset includes images with various page styles like ruled lines, gridlines, and partially blank pages that are evident in our day-to-day encounters. Further, they lack the document images with realistic degradations like crumpled pages, non-uniform illumination, and shadow gradients that are well incorporated in the proposed LS-HDIB dataset. These attributes enhance the diversity and the versatility of the proposed dataset.

\subsection{Training Details} 
\label{sec:train_details}

We use eight widely used deep networks - DeepLabV3 \cite{deeplabv3}, DeepLabV3+ \cite{deeplabv3plus}, Feature Pyramid Networks (FPN) \cite{fpnetwork}, LinkNet \cite{linknet}, Multi-scale Attention net (MANet) \cite{manet}, Pyramid Attention Network (PAN) \cite{panet}, Pyramid Scene Parsing Network (PSPN) \cite{pascenenet}, and U-Net \cite{unet} - to understand the effectiveness of the proposed dataset for the handwritten document image binarization task. We have carefully chosen these networks as they collectively span the various deep learning based approaches \cite{tensmeyer2019generating, peng2017using, calvo2017pixel, improvegan} that have been adopted thus far for the binarization task. 

Each of the eight networks is trained under three different training regimes. We follow the standard train, validation, and test split of $80\%$, $10\%$, and $10\%$, respectively, for each regime.\\
(i) \textbf{Regime 1}: The deep models are trained only on the \textit{baseline dataset} obtained by combining ten different publicly available datasets: DIBCO09 \cite{dibco9}, HDIBCO10 \cite{hdibco10}, DIBCO11 \cite{dibco11}, HDIBCO12\cite{hdibco12}, DIBCO13 \cite{dibco13}, HDIBCO14 \cite{hdibco14}, PHIBD12 \cite{phdib}, HDIBCO16 \cite{hdibco16}, DIBCO17 \cite{dibco17}, DIBCO18 \cite{dibco18}. However, even after combining ten different datasets, the size of the baseline dataset is relatively small (order of 1000 images). Therefore, we crop the full-length document images of the baseline dataset to the size $480 \times 480$ with a stride of $240$ pixels and perform rotation ($90^\circ, 180^\circ$, and $270^\circ$) and horizontal flip to obtain around $6000$ images, $5000$ for training and $1000$ for testing. While the size of the baseline dataset can further be increased by reducing the stride value, this leads to greater overlap and high redundancy in the foreground content across different content images.\\
(ii) \textbf{Regime 2}: Each deep model is trained only on the LS-HDIB training set. Although the LS-HDIB dataset has over 1 million images, we use only $5000$ images for training to have a fair performance comparison.\\ 
(iii) \textbf{Regime 3}: We combine both the LS-HDIB and the baseline dataset, by randomly selecting $2500$ images from each dataset to train all the deep models on a total of $5000$ images. Regime 3 is targeted towards establishing the efficacy of augmenting the proposed dataset to the existing ones.

% Each of the eight networks is trained under three different training regimes. We follow the standard train, validation, and test split of $80\%$, $10\%$, and $10\%$, respectively, for each regime.\\
% (i) \textbf{Regime 1}: The deep models are trained only on the \textit{baseline dataset} obtained by combining ten different publicly available datasets: DIBCO09, HDIBCO10, DIBCO11, HDIBCO12, DIBCO13, HDIBCO14, PHIBD12, HDIBCO16, DIBCO17, DIBCO18 \cite{dibco18}. However, even after combining ten different datasets, the size of the baseline dataset is relatively small (order of 1000 images). Therefore, we crop the full-length document images of the baseline dataset to the size $480 \times 480$ with a stride of $240$ pixels and perform rotation ($90^\circ, 180^\circ$, and $270^\circ$) and horizontal flip to obtain around $5000$ images for training. While the size of the baseline dataset can further be increased by reducing the stride value, this leads to greater overlap and high redundancy in the foreground content across different content images.\\
% (ii) \textbf{Regime 2}: Each of the deep model is trained only on the LS-HDIB training set. Although the LS-HDIB dataset has over 1 million images, we use only $5000$ images for training to have a fair performance comparison.\\ 
% (iii) \textbf{Regime 3}: We combine both the LS-HDIB and the baseline dataset, by randomly selecting $2500$ images from each dataset to train all the deep models on a total of $5000$ images. Regime 3 is targeted towards establishing the efficacy of augmenting the proposed dataset to the existing ones.

The models are trained for a maximum of $20$ epochs with learning rate of $0.01$ using the Adam optimizer with default parameters.  The training is performed on the NVIDIA RTX 2080 Ti GPU with the batch size of $8$.\\
\\
\textbf{Loss Function.} We use binary cross-entropy loss to train the segmentation models, as described in Equation \ref{eq:4}. 
\begin{equation}\label{eq:4}
\mathcal{L} = - I_{gt} \mathrm{log}\left(\mathcal{F}(I_{in})\right) - (1 - I_{gt}) \mathrm{log}\left(1 - \mathcal{F}(I_{in})\right)    
\end{equation}
Here, $\mathcal{F}$ represents the functional form of deep model. Binary cross-entropy loss is found to be more effective than MSE loss for classification tasks \cite{crossvsmse}.

\begin{figure*}[ht]
    \centering
    \includegraphics[width=\textwidth]{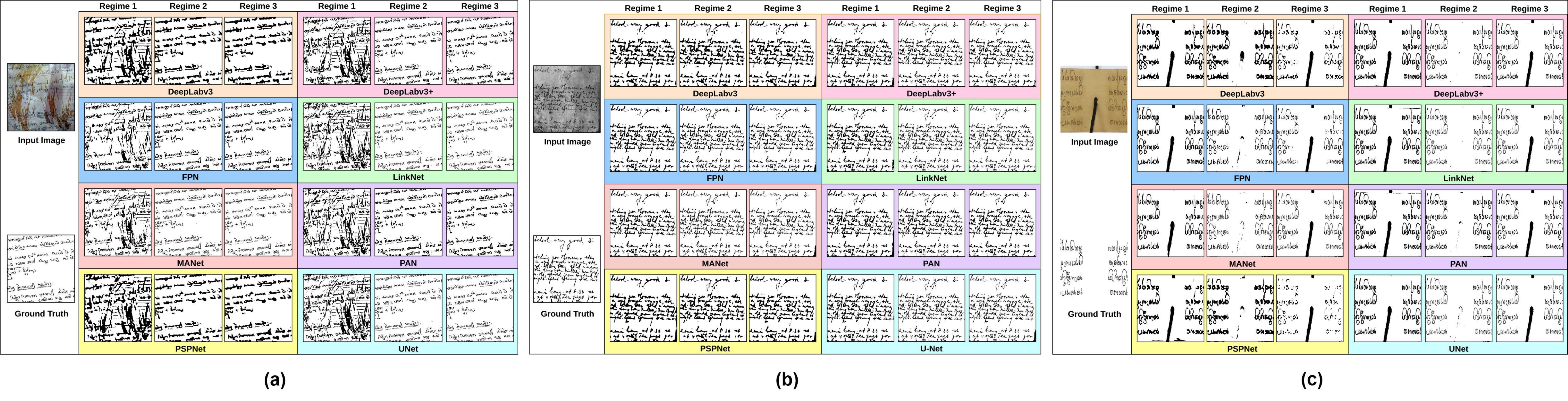}
    \caption{Qualitative result on (a) the LS-HDIB test set (b) Bickley Diary dataset, and (c) Palm Leaf Manuscript dataset.}
    \label{fig:qual_res}
    \vspace{-4mm}
\end{figure*}

\section{Experimental Analysis}
\label{sec:experiments}
We demonstrate the effectiveness of the proposed dataset for handwritten document image binarization through an extensive quantitative and qualitative analysis. We compare the performance of different deep models trained to observe how well the LS-HDIB dataset enhances the generalization capability of the deep models.

We use four different datasets containing challenging scenarios for testing the deep models: the Bickley Diary dataset \cite{bickleydiary}, the Palm Leaf Manuscript dataset \cite{palmleafmanuscript}, the DIBCO test set, and the LS-HDIB test set. We use three popular metrics to evaluate the network performance trained under different regimes: F-measure $(\mathrm{F_{score}})$ \cite{evalmet}, pseudo F-measure $(\mathrm{PF_{score}})$ \cite{evalmet}, and Peak Signal to Noise Ratio $(\mathrm{PSNR})$ \cite{evalmet}.

As shown in Table \ref{table:quant_res}, the $\mathrm{F_{score}}$, $\mathrm{PF_{score}}$, and $\mathrm{PSNR}$ is maximum over all the deep models trained under Regime 2 for LS-HDIB and the Bickley Diary dataset. Further, the performance under Regime 3 is better than that of Regime 1, indicating that augmenting the LS-HDIB dataset to the baseline dataset enhances network performance. Qualitatively, the foreground content of the image affected by liquid stains, noisy background, and poor foreground to background contrast is well recovered under Regime 2, as shown in Fig. \ref{fig:qual_res}(a) and \ref{fig:qual_res}(b). For the Palm Leaf dataset, the strokes in the estimated foreground content corresponding to Regime 1 (and 3) are relatively thicker when compared to those corresponding to Regime 2 across all the models (see Fig. \ref{fig:qual_res}(c)). Since the stroke widths are consistent with the ground truth, the $\mathrm{PSNR}$ continues to be higher for Regime 2. However, the $\mathrm{F_{score}}$ and $\mathrm{PF_{score}}$ of some models under Regime 2 and 3 are less than Regime 1. This is attributed to the presence of minor discontinuities in the strokes obtained under Regime 2 when compared to that of Regime 1 (see Fig. \ref{fig:qual_res}(c)). However, on average, the overall performance of each of the eight deep models is the highest when trained under Regime 2 across the three different test datasets, as evident from Table \ref{table:quant_res}. Further, Fig. \ref{fig:qual_res}(c) depicts that models under Regime 1 and 3 fail to segment out the thread punched out through the document. However, almost every deep model trained on the LS-HDIB dataset (Regime 2) precisely identifies the appropriate foreground content even in the presence of such background artifacts. 

For the DIBCO test set, the performance under Regime 1 is better than Regime 2 (Table \ref{table:quant_res}). This is inline with the expectation as the networks are trained on DIBCO train set itself. Interestingly, Regime 3 offers the best performance across different models and test sets indicating that the LS-HDIB dataset when augmented with the standard DIBCO datasets enhances the network performance. 

\begin{table}[h]
\centering
\resizebox{\linewidth}{!}{%
\begin{tabular}{l|l|cccccccc|}
\hline
\multicolumn{1}{c|}{Metric} & \multicolumn{1}{c|}{Dataset} & DeepLabv3 & DeepLabv3++ & FPN & LinkNet & MANet & PANet & PSPNet & U-Net \\ \hline
 &  & {\color[HTML]{00009B} 0.64} & {\color[HTML]{00009B} 0.71} & {\color[HTML]{00009B} 0.72} & {\color[HTML]{00009B} 0.71} & {\color[HTML]{00009B} 0.72} & {\color[HTML]{00009B} 0.72} & {\color[HTML]{00009B} 0.58} & {\color[HTML]{00009B} 0.74} \\
 &  & {\color[HTML]{FE0000} \textbf{0.66}} & {\color[HTML]{FE0000} \textbf{0.77}} & {\color[HTML]{FE0000} \textbf{0.77}} & {\color[HTML]{FE0000} \textbf{0.82}} & {\color[HTML]{FE0000} \textbf{0.84}} & {\color[HTML]{FE0000} \textbf{0.74}} & {\color[HTML]{FE0000} \textbf{0.65}} & {\color[HTML]{FE0000} \textbf{0.83}} \\
 & \multirow{-3}{*}{Bickley Diary} & {\color[HTML]{036400} 0.60} & {\color[HTML]{036400} 0.75} & {\color[HTML]{036400} 0.74} & {\color[HTML]{036400} 0.80} & {\color[HTML]{036400} 0.82} & {\color[HTML]{036400} 0.69} & {\color[HTML]{036400} 0.60} & {\color[HTML]{036400} 0.77} \\ \cline{2-10}
 &  & {\color[HTML]{00009B} \textbf{0.49}} & {\color[HTML]{00009B} \textbf{0.59}} & {\color[HTML]{00009B} 0.58} & {\color[HTML]{00009B} \textbf{0.62}} & {\color[HTML]{00009B} \textbf{0.62}} & {\color[HTML]{00009B} \textbf{0.57}} & {\color[HTML]{00009B} 0.47} & {\color[HTML]{00009B} \textbf{0.62}} \\
 &  & {\color[HTML]{FE0000} 0.48} & {\color[HTML]{FE0000} 0.57} & {\color[HTML]{FE0000} \textbf{0.59}} & {\color[HTML]{FE0000} 0.60} & {\color[HTML]{FE0000} \textbf{0.62}} & {\color[HTML]{FE0000} \textbf{0.57}} & {\color[HTML]{FE0000} \textbf{0.48}} & {\color[HTML]{FE0000} 0.60} \\
 & \multirow{-3}{*}{Palm Leaf} & {\color[HTML]{036400} 0.48} & {\color[HTML]{036400} 0.58} & {\color[HTML]{036400} 0.57} & {\color[HTML]{036400} 0.61} & {\color[HTML]{036400} \textbf{0.62}} & {\color[HTML]{036400} \textbf{0.57}} & {\color[HTML]{036400} 0.44} & {\color[HTML]{036400} 0.61} \\ \cline{2-10}
 &  & {\color[HTML]{00009B} 0.48} & {\color[HTML]{00009B} \textbf{0.68}} & {\color[HTML]{00009B} 0.60} & {\color[HTML]{00009B} 0.66} & {\color[HTML]{00009B} 0.68} & {\color[HTML]{00009B} 0.56} & {\color[HTML]{00009B} \textbf{0.50}} & {\color[HTML]{00009B} 0.67} \\
 &  & {\color[HTML]{FE0000} 0.43} & {\color[HTML]{FE0000} 0.57} & {\color[HTML]{FE0000} 0.46} & {\color[HTML]{FE0000} 0.58} & {\color[HTML]{FE0000} 0.60} & {\color[HTML]{FE0000} 0.41} & {\color[HTML]{FE0000} 0.43} & {\color[HTML]{FE0000} 0.58} \\
 & \multirow{-3}{*}{DIBCO} & {\color[HTML]{036400} \textbf{0.52}} & {\color[HTML]{036400} 0.62} & {\color[HTML]{036400} \textbf{0.61}} & {\color[HTML]{036400} \textbf{0.72}} & {\color[HTML]{036400} \textbf{0.71}} & {\color[HTML]{036400} \textbf{0.58}} & {\color[HTML]{036400} 0.46} & {\color[HTML]{036400} \textbf{0.72}} \\ \cline{2-10}
 &  & {\color[HTML]{00009B} 0.44} & {\color[HTML]{00009B} 0.52} & {\color[HTML]{00009B} 0.54} & {\color[HTML]{00009B} 0.56} & {\color[HTML]{00009B} 0.57} & {\color[HTML]{00009B} 0.51} & {\color[HTML]{00009B} 0.42} & {\color[HTML]{00009B} 0.55} \\
 &  & {\color[HTML]{FE0000} \textbf{0.54}} & {\color[HTML]{FE0000} \textbf{0.68}} & {\color[HTML]{FE0000} 0.68} & {\color[HTML]{FE0000} \textbf{0.86}} & {\color[HTML]{FE0000} \textbf{0.87}} & {\color[HTML]{FE0000} \textbf{0.65}} & {\color[HTML]{FE0000} \textbf{0.53}} & {\color[HTML]{FE0000} \textbf{0.87}} \\
\multirow{-12}{*}{F-score} & \multirow{-3}{*}{LS-HDIB} & {\color[HTML]{036400} 0.53} & {\color[HTML]{036400} 0.66} & {\color[HTML]{036400} 0.67} & {\color[HTML]{036400} 0.85} & {\color[HTML]{036400} 0.85} & {\color[HTML]{036400} 0.64} & {\color[HTML]{036400} 0.52} & {\color[HTML]{036400} 0.86} \\ \hline
 &  & {\color[HTML]{00009B} 0.64} & {\color[HTML]{00009B} 0.76} & {\color[HTML]{00009B} 0.75} & {\color[HTML]{00009B} 0.80} & {\color[HTML]{00009B} 0.79} & {\color[HTML]{00009B} 0.71} & {\color[HTML]{00009B} 0.59} & {\color[HTML]{00009B} 0.78} \\
 &  & {\color[HTML]{FE0000} \textbf{0.66}} & {\color[HTML]{FE0000} \textbf{0.79}} & {\color[HTML]{FE0000} \textbf{0.79}} & {\color[HTML]{FE0000} \textbf{0.87}} & {\color[HTML]{FE0000} \textbf{0.89}} & {\color[HTML]{FE0000} \textbf{0.76}} & {\color[HTML]{FE0000} \textbf{0.66}} & {\color[HTML]{FE0000} \textbf{0.89}} \\
 & \multirow{-3}{*}{Bickley Diary} & {\color[HTML]{036400} 0.60} & {\color[HTML]{036400} 0.77} & {\color[HTML]{036400} 0.75} & {\color[HTML]{036400} 0.82} & {\color[HTML]{036400} 0.80} & {\color[HTML]{036400} 0.73} & {\color[HTML]{036400} 0.61} & {\color[HTML]{036400} 0.82} \\ \cline{2-10}
 &  & {\color[HTML]{00009B} \textbf{0.50}} & {\color[HTML]{00009B} \textbf{0.60}} & {\color[HTML]{00009B} 0.57} & {\color[HTML]{00009B} 0.61} & {\color[HTML]{00009B} 0.62} & {\color[HTML]{00009B} \textbf{0.58}} & {\color[HTML]{00009B} \textbf{0.48}} & {\color[HTML]{00009B} \textbf{0.63}} \\
 &  & {\color[HTML]{FE0000} 0.48} & {\color[HTML]{FE0000} 0.58} & {\color[HTML]{FE0000} \textbf{0.59}} & {\color[HTML]{FE0000} \textbf{0.63}} & {\color[HTML]{FE0000} \textbf{0.64}} & {\color[HTML]{FE0000} 0.57} & {\color[HTML]{FE0000} \textbf{0.48}} & {\color[HTML]{FE0000} 0.62} \\
 & \multirow{-3}{*}{Palm Leaf} & {\color[HTML]{036400} 0.48} & {\color[HTML]{036400} 0.59} & {\color[HTML]{036400} 0.58} & {\color[HTML]{036400} 0.62} & {\color[HTML]{036400} \textbf{0.64}} & {\color[HTML]{036400} 0.57} & {\color[HTML]{036400} 0.44} & {\color[HTML]{036400} \textbf{0.63}} \\\cline{2-10}
 &  & {\color[HTML]{00009B} 0.49} & {\color[HTML]{00009B} \textbf{0.68}} & {\color[HTML]{00009B} 0.60} & {\color[HTML]{00009B} 0.66} & {\color[HTML]{00009B} 0.68} & {\color[HTML]{00009B} 0.56} & {\color[HTML]{00009B} \textbf{0.50}} & {\color[HTML]{00009B} 0.67} \\
 &  & {\color[HTML]{FE0000} 0.41} & {\color[HTML]{FE0000} 0.52} & {\color[HTML]{FE0000} 0.51} & {\color[HTML]{FE0000} 0.58} & {\color[HTML]{FE0000} 0.51} & {\color[HTML]{FE0000} 0.41} & {\color[HTML]{FE0000} 0.43} & {\color[HTML]{FE0000} 0.58} \\
 & \multirow{-3}{*}{DIBCO} & {\color[HTML]{036400} \textbf{0.52}} & {\color[HTML]{036400} 0.62} & {\color[HTML]{036400} \textbf{0.62}} & {\color[HTML]{036400} \textbf{0.73}} & {\color[HTML]{036400} \textbf{0.72}} & {\color[HTML]{036400} \textbf{0.59}} & {\color[HTML]{036400} 0.47} & {\color[HTML]{036400} \textbf{0.72}} \\\cline{2-10}
 &  & {\color[HTML]{00009B} 0.43} & {\color[HTML]{00009B} 0.51} & {\color[HTML]{00009B} 0.53} & {\color[HTML]{00009B} 0.56} & {\color[HTML]{00009B} 0.56} & {\color[HTML]{00009B} 0.51} & {\color[HTML]{00009B} 0.42} & {\color[HTML]{00009B} 0.54} \\
 &  & {\color[HTML]{FE0000} \textbf{0.53}} & {\color[HTML]{FE0000} \textbf{0.67}} & {\color[HTML]{FE0000} \textbf{0.66}} & {\color[HTML]{FE0000} \textbf{0.87}} & {\color[HTML]{FE0000} \textbf{0.88}} & {\color[HTML]{FE0000} \textbf{0.64}} & {\color[HTML]{FE0000} \textbf{0.52}} & {\color[HTML]{FE0000} \textbf{0.88}} \\
\multirow{-12}{*}{PF-score} & \multirow{-3}{*}{LS-HDIB} & {\color[HTML]{036400} 0.52} & {\color[HTML]{036400} 0.65} & {\color[HTML]{036400} \textbf{0.66}} & {\color[HTML]{036400} 0.86} & {\color[HTML]{036400} 0.86} & {\color[HTML]{036400} 0.63} & {\color[HTML]{036400} 0.52} & {\color[HTML]{036400} 0.87} \\ \hline
 &  & {\color[HTML]{00009B} \textbf{10.07}} & {\color[HTML]{00009B} 12.32} & {\color[HTML]{00009B} 12.11} & {\color[HTML]{00009B} 13.16} & {\color[HTML]{00009B} 12.83} & {\color[HTML]{00009B} 11.55} & {\color[HTML]{00009B} 9.61} & {\color[HTML]{00009B} 12.89} \\
 &  & {\color[HTML]{FE0000} 10.06} & {\color[HTML]{FE0000} \textbf{12.75}} & {\color[HTML]{FE0000} \textbf{12.69}} & {\color[HTML]{FE0000} \textbf{14.42}} & {\color[HTML]{FE0000} \textbf{14.82}} & {\color[HTML]{FE0000} \textbf{12.10}} & {\color[HTML]{FE0000} \textbf{10.01}} & {\color[HTML]{FE0000} \textbf{14.64}} \\
 & \multirow{-3}{*}{Bickley Diary} & {\color[HTML]{036400} 9.69} & {\color[HTML]{036400} 12.40} & {\color[HTML]{036400} 12.29} & {\color[HTML]{036400} 13.41} & {\color[HTML]{036400} 13.49} & {\color[HTML]{036400} 11.64} & {\color[HTML]{036400} 9.78} & {\color[HTML]{036400} 13.24} \\\cline{2-10}
 &  & {\color[HTML]{00009B} 9.02} & {\color[HTML]{00009B} 10.53} & {\color[HTML]{00009B} 10.68} & {\color[HTML]{00009B} 11.04} & {\color[HTML]{00009B} 11.27} & {\color[HTML]{00009B} 10.41} & {\color[HTML]{00009B} 9.04} & {\color[HTML]{00009B} 11.11} \\
 &  & {\color[HTML]{FE0000} 8.93} & {\color[HTML]{FE0000} \textbf{11.57}} & {\color[HTML]{FE0000} \textbf{11.47}} & {\color[HTML]{FE0000} \textbf{12.45}} & {\color[HTML]{FE0000} \textbf{12.31}} & {\color[HTML]{FE0000} \textbf{11.12}} & {\color[HTML]{FE0000} 8.99} & {\color[HTML]{FE0000} \textbf{12.41}} \\
 & \multirow{-3}{*}{Palm Leaf} & {\color[HTML]{036400} \textbf{9.19}} & {\color[HTML]{036400} 11.34} & {\color[HTML]{036400} 11.36} & {\color[HTML]{036400} 12.26} & {\color[HTML]{036400} 12.24} & {\color[HTML]{036400} 11.03} & {\color[HTML]{036400} \textbf{9.36}} & {\color[HTML]{036400} 12.32} \\\cline{2-10}
 &  & {\color[HTML]{00009B} 8.26} & {\color[HTML]{00009B} 9.76} & {\color[HTML]{00009B} 9.63} & {\color[HTML]{00009B} 10.84} & {\color[HTML]{00009B} 11.13} & {\color[HTML]{00009B} 9.01} & {\color[HTML]{00009B} 7.90} & {\color[HTML]{00009B} 11.06} \\
 &  & {\color[HTML]{FE0000} 6.75} & {\color[HTML]{FE0000} 8.97} & {\color[HTML]{FE0000} 8.74} & {\color[HTML]{FE0000} 10.96} & {\color[HTML]{FE0000} 9.54} & {\color[HTML]{FE0000} \textbf{10.53}} & {\color[HTML]{FE0000} 6.56} & {\color[HTML]{FE0000} 11.01} \\
 & \multirow{-3}{*}{DIBCO} & {\color[HTML]{036400} \textbf{8.38}} & {\color[HTML]{036400} \textbf{10.47}} & {\color[HTML]{036400} \textbf{10.34}} & {\color[HTML]{036400} \textbf{12.54}} & {\color[HTML]{036400} \textbf{12.36}} & {\color[HTML]{036400} 9.74} & {\color[HTML]{036400} \textbf{8.39}} & {\color[HTML]{036400} \textbf{12.56}} \\\cline{2-10}
 &  & {\color[HTML]{00009B} 7.58} & {\color[HTML]{00009B} 9.29} & {\color[HTML]{00009B} 9.83} & {\color[HTML]{00009B} 9.63} & {\color[HTML]{00009B} 9.86} & {\color[HTML]{00009B} 9.26} & {\color[HTML]{00009B} 7.49} & {\color[HTML]{00009B} 9.05} \\
 &  & {\color[HTML]{FE0000} \textbf{9.39}} & {\color[HTML]{FE0000} \textbf{12.15}} & {\color[HTML]{FE0000} 12.05} & {\color[HTML]{FE0000} \textbf{16.99}} & {\color[HTML]{FE0000} \textbf{17.09}} & {\color[HTML]{FE0000} \textbf{11.58}} & {\color[HTML]{FE0000} \textbf{9.37}} & {\color[HTML]{FE0000} \textbf{17.29}} \\
\multirow{-12}{*}{PSNR} & \multirow{-3}{*}{LS-HDIB} & {\color[HTML]{036400} 9.11} & {\color[HTML]{036400} 11.88} & {\color[HTML]{036400} \textbf{12.08}} & {\color[HTML]{036400} 16.63} & {\color[HTML]{036400} 16.74} & {\color[HTML]{036400} 11.33} & {\color[HTML]{036400} 9.10} & {\color[HTML]{036400} 17.07}\\ \hline
\end{tabular}%
}

\caption{F\textsubscript{score}, PF\textsubscript{score}, and PSNR evaluated over eight deep models under three different regimes: {\color[HTML]{00009B}Regime 1}, {\color[HTML]{FE0000}Regime 2}, and {\color[HTML]{036400} Regime 3} over test datasets.}
\label{table:quant_res}
\end{table}

\begin{figure}[h]
    \centering
    \includegraphics[width=\linewidth]{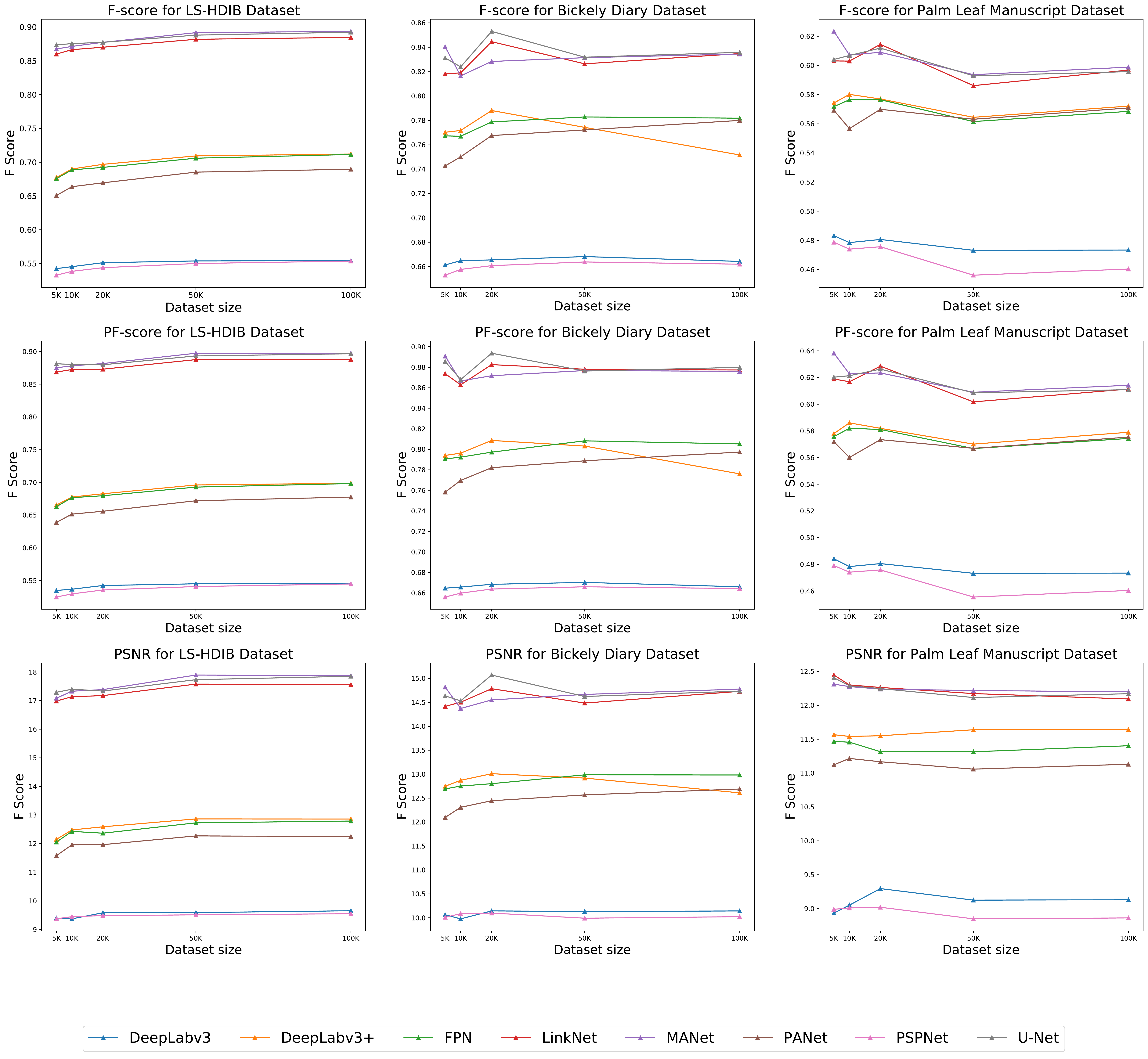}
    \caption{Effect of varying dataset size on the model performance evaluated over the three test datasets.}
    \label{fig:test_trend}
    \vspace{-4mm}
\end{figure}

Given that nearly all the models have performed the best when trained on LS-HDIB dataset across all the three test datasets, we further investigate the effect of dataset size on the network performance. For LS-HDIB test dataset, the $\mathrm{F_{score}}$, $\mathrm{PF_{score}}$, and $\mathrm{PSNR}$ are observed to increase with the dataset size, as shown in Fig. \ref{fig:test_trend}. This indicates the requirement of a large-scale dataset to span complex real-world scenarios encountered in the handwritten document images. Fig. \ref{fig:test_trend} shows that the performance over the Bickley Diary and Palm Leaf dataset peaks at the dataset size of $20$K and $5$K, respectively. Overall, we have established that the deep models are more robust to various degradation effects and page styles encountered in the real world when trained on the LS-HDIB dataset and necessitate the need for such a dataset.

\textbf{Note.} Owing to the space constraints, we provide more qualitative results, detailed dataset statistics, training and validation logs for different models (for better selectivity), and the accompanying code on our website\footnote{\textcolor{blue}{\url{https://kaustubh-sadekar.github.io/LS-HDIB/}}}.

\section{Conclusion}
\label{sec:conclusion}

We propose a large-scale dataset (LS-HDIB) for handwritten document image binarization and a simple yet effective method to generate it. When trained on the LS-HDIB dataset, different deep models can generalize better on unseen document images with a wide variety of degradations encountered in our day-to-day lives. This is possible due to the inherent diversity of the proposed dataset. Further, this work highlights that the fundamental image processing algorithms can be used as practical tools to support the existing deep-learning-based methods in producing significantly better results. In our case, we use adaptive thresholding and mixed gradient-based seamless cloning to generate this large-scale dataset.

\newpage
% -------------------------------------------------------------------------
\bibliographystyle{IEEEbib}
{\footnotesize\bibliography{strings,refs}}

\end{document}